\documentclass[10pt,twocolumn,letterpaper]{article}

\usepackage{cvpr}              % To produce the CAMERA-READY version
\usepackage{float}
\usepackage{stfloats}
\usepackage{bbding}
\usepackage{xcolor}
\usepackage{subcaption}
\usepackage{multirow}
\usepackage{threeparttable}
\usepackage[normalem]{ulem}
\definecolor{cvprblue}{rgb}{0.21,0.49,0.74}
\definecolor{myorange}{rgb}{1, 0.85, 0.7}
\definecolor{myred}{rgb}{1, 0.7, 0.7}

\newcommand{\reducedstrut}{\vrule width 0pt height 1.05\ht\strutbox depth 1.0\dp\strutbox\relax}
\newcommand{\sota}[1]{%
  \begingroup
  \setlength{\fboxsep}{0pt}%  
  \colorbox{myred}{\reducedstrut#1\/}%
  \endgroup
}
\newcommand{\subsota}[1]{%
  \begingroup
  \setlength{\fboxsep}{0pt}%  
  \colorbox{myorange}{\reducedstrut#1\/}%
  \endgroup
}
\newcommand{\revise}[1]{#1}

\newcommand{\del}[1]{}
\newcommand{\added}[1]{#1}

\newcommand{\yx}[1]{{#1}}
\newcommand{\zyx}[1]{{#1}}

\usepackage[pagebackref,breaklinks,colorlinks,allcolors=cvprblue]{hyperref}

\def\paperID{} % *** Enter the Paper ID here
\def\confName{CVPR}
\def\confYear{2026}

\title{\zyx{IntrinsicWeather: Controllable Weather Editing in Intrinsic Space}}

\author{
Yixin Zhu\textsuperscript{1} \quad
Zuo-Liang Zhu\textsuperscript{2} \quad
Jian Yang\textsuperscript{1} \quad
Miloš Hašan\textsuperscript{3} \quad
Jin Xie\textsuperscript{1*} \quad
Beibei Wang\textsuperscript{1*} \\[2pt] 
{
\fontsize{10pt}{12pt}\selectfont 
\textsuperscript{1}Nanjing University \quad
\textsuperscript{2}Nankai University \quad
\textsuperscript{3}NVIDIA
}
}

\begin{document}

\twocolumn[{%
\renewcommand\twocolumn[1][]{#1}%
\maketitle
\includegraphics[width=1.0\linewidth]{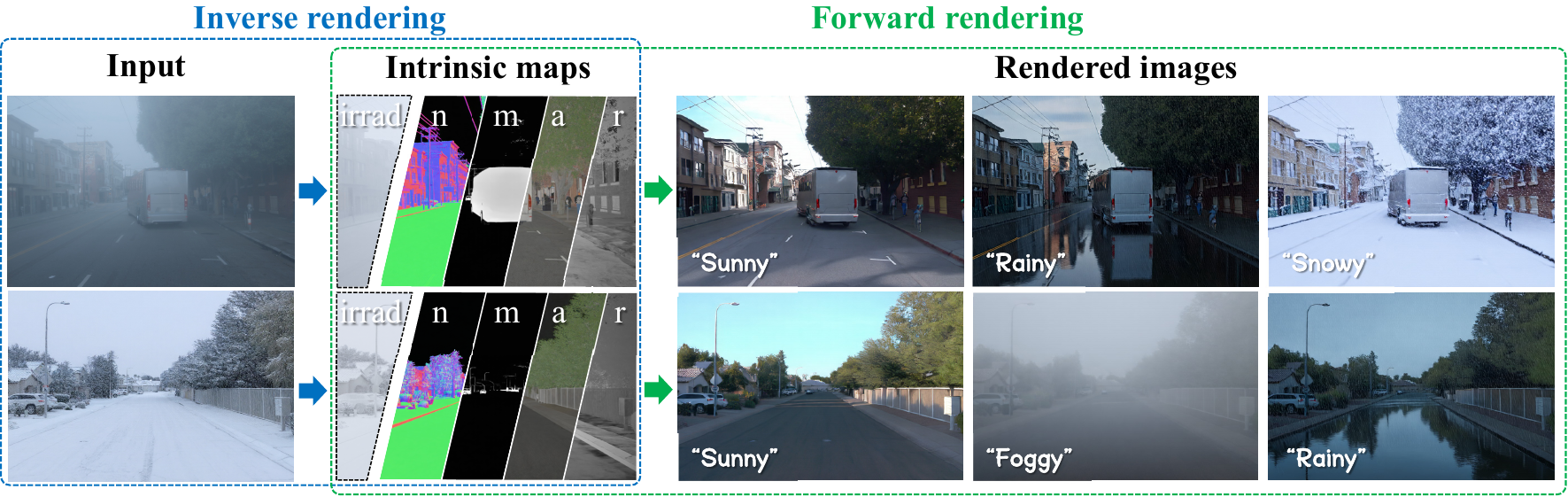}
% \vspace{-2em}
\captionof{figure}{\added{We introduce IntrinsicWeather, a framework for controllable weather editing in intrinsic space. Our framework includes two components, an inverse renderer and a forward renderer.
The inverse renderer decomposes an input image into intrinsic maps, including weather-invariant material maps (albedo, roughness, metallicity), a normal map, and an irradiance map that captures illumination and weather effects.
The forward renderer then combines these maps with a prompt specifying the target weather to synthesize a new image.
By disentangling materials, geometry, and illumination, IntrinsicWeather enables realistic and controllable weather manipulation.
}}
\label{fig:teaser}
\vspace{5mm}
}]
{
  \renewcommand{\thefootnote}{}
  \footnotetext{\textsuperscript{*} Corresponding author.}
  \footnotetext{\textsuperscript{1} School of Intelligence Science and Technology, Nanjing University, Suzhou, China}
  \footnotetext{\textsuperscript{2} College of Computer Science, Nankai University, Tianjin, China}
}

\begin{abstract}
% Forward and inverse rendering have emerged as key techniques for enabling understanding and reconstruction in the context of autonomous driving (AD). However, complex weather and illumination pose great challenges to this task. The emergence of large diffusion models has shown promise in achieving reasonable results through learning from 2D priors, but these models are difficult to control and lack robustness. 
% We present IntrinsicWeather, a diffusion-based framework for forward and inverse rendering of autonomous driving scenes under various weather and lighting conditions. 
\added{We present IntrinsicWeather, a diffusion-based framework for controllable weather editing in intrinsic space.} Our framework includes two components based on diffusion priors: an inverse renderer that estimates material properties, scene geometry, and lighting as intrinsic maps from an input image, and a forward renderer that utilizes these geometry and material maps along with a text prompt that describes specific weather conditions to generate a final image. The intrinsic maps enhance controllability compared to traditional pixel-space editing approaches.
%Several challenges for building such a framework include ensuring the spatial correspondence between predicted intrinsic maps and image regions during generation, particularly for outdoor scenes, as well as addressing dataset scarcity. To overcome these challenges,
We propose an intrinsic map-aware attention mechanism that improves spatial correspondence and decomposition quality in large outdoor scenes. For forward rendering, we leverage CLIP-space interpolation of weather prompts to achieve fine-grained weather control. \added{We also introduce a synthetic and a real-world dataset, containing 38k and 18k images under various weather conditions, each with intrinsic map annotations.} \added{IntrinsicWeather outperforms state-of-the-art pixel-space editing approaches, weather restoration methods, and rendering-based methods}, showing promise for downstream tasks such as autonomous driving, enhancing the robustness of detection and segmentation in challenging weather scenarios. This project is available at \url{https://yixinzhu042.github.io/IntrinsicWeather/}.

%Our inverse rendering model enables estimation of material properties, scene geometry, and lighting, while the forward rendering model supports controllable weather and illumination editing through the use of predicted intrinsic maps guided by text descriptions.

%\new{For inverse rendering, we propose intrinsic map-aware attention that enforces spatial correspondence between predicted intrinsic maps and image regions, improving decomposition quality on large outdoor scenes. For forward rendering, we leverage CLIP-space interpolation of weather prompts and SD-priors to achieve fine-grained weather control.}
% Yet, we find that conventional diffusion models overlook the spatial correspondence between intrinsic maps and image regions, leading to a lack of attention to important areas. We observe that different intrinsic maps should correspond to different regions of the original image. Based on this observation, we extend diffusion with intrinsic map-aware attention to enable high-quality inverse rendering. 
%We also introduce two datasets (\ie, WeatherSynthetic (38k) and WeatherReal (18k)) with intrinsic maps annotations. Extensive experiments show that our IntrinsicWeather outperforms state-of-the-art methods on several benchmarks. Moreover, our method demonstrates significant value in downstream tasks for autonomous driving, enhancing the robustness of object detection and \revise{semantic segmentation} in challenging weather scenarios.
\end{abstract}   
%------------------------------------%
\section{Introduction}
\label{sec:intro}
%------------------------------------%

%\added{besides the visual comparison, do we need to compare ours vs weatherweaver on these two task? }

For autonomous vehicles, robust scene understanding requires the ability to handle adverse weather conditions.
% One promising approach is to use simulated weather-conditioned data to improve a perception model's generalization to corner cases.
\added{Techniques that handle weather effects, can substantially improve the robustness of perception models across different weather conditions.}
While diffusion-based image editing techniques~\cite{labs2025flux1kontextflowmatching, wu2025qwenimagetechnicalreport, brooks2023instructpix2pix, hertz2022prompttopromptimageeditingcross} have created significant opportunities here, a key limitation remains: the lack of fine-grained controllability in the generated scenarios.

% Existing approaches weather editing approaches?.  Unified one.  Particularly, Lin et al..... However, these methods....

\added{
% Existing editing approaches apply editing in pixel space, using a unified diffusion model~\cite{ho2020denoising,rombach2022high,ramesh2022hierarchical,esser2024scaling} to transform the input into a target image following the description. 
Following existing image editing approaches, weather editing methods~\cite{cosne2020using, li2021weatherganmultidomainweather, tasnim2025normalizing} typically leverage a unified generative model to perform weather transformations in the pixel space. Particularly, WeatherWeaver~\cite{lin2025controllable} decomposes weather editing into a two-stage strategy: weather removal and weather synthesis. While these methods have achieved impressive results in weather manipulation, they still struggle to preserve the underlying material and geometry of the scene while generating natural illumination.
}

% \added{do we need to mention intrinsic space object manipulation here?}

In this paper, we present \emph{IntrinsicWeather}, a novel framework for \added{controllable weather editing} that operates \added{in intrinsic space.} Our key insight is that by decomposing a scene into its fundamental components—material properties, geometry, and lighting, we can achieve higher control over weather effects than in the image space. IntrinsicWeather consists of two key components: a forward renderer that uses CLIP-space interpolation and diffusion priors for fine-grained weather synthesis, and a novel inverse renderer \yx{tailored for outdoor and autonomous driving scenes.} 
% \milos{Do we really want to sell autonomous driving so much? The reviewers may tell us it is too niche. How about we say "outdoor and autonomous driving scenes" to make it more general?} 
We are inspired by recent diffusion-based intrinsic decomposition and recomposition approaches, RGB$\leftrightarrow$X~\cite{zeng2024rgb} and DiffusionRenderer \cite{liang2025diffusionrenderer}, which have achieved impressive results at the indoor and object levels; however, they do not focus on weather and do not generalize to large-scale autonomous driving scenarios.  We bridge this gap with a novel intrinsic map-aware attention mechanism that ensures spatial correspondence, thereby significantly enhancing decomposition fidelity in complex, unconstrained outdoor environments. \added{By combining the inverse and forward renderers, our framework allows for controlled editing of weather and lighting.}

%our key idea is to perform the \added{weather controled generation} in the intrinsic space, instead of the image space. In the intrinsic space, the color has been decomposed into material properties, scene geometry, and lighting, leading to more controble generation. Based on this key insight, we introduce 
%the current existing intrinsic estimation and rendering approaches, \added{such as xxx, and xx}, are trained \added{on indoor scenes and without any weather conditions.} They can not be leveraged for our IntrinsicWeather directly, as the distribution of the scale is more complex, raising difficulties in generation.

% \added{We also introduce two datasets (\ie, WeatherSynthetic (38k) and WeatherReal (18k)) with intrinsic maps annotations to train these two components.} 
 
% \added{Our IntrinsicWeather outperforms existing state-of-the-art inverse and forward rendering methods, weather restoration methods, and pixel-space editing methods, achieving higher-quality inverse rendering results and more natural and controllable weather editing results.} 
\added{Our IntrinsicWeather outperforms existing state-of-the-art inverse and forward rendering methods, achieving over 10 dB PSNR improvement in inverse rendering and higher PickScore~\cite{kirstain2023pickapicopendatasetuser} in forward rendering. Compared with weather restoration and pixel-space editing methods, our approach achieves higher text-image consistency and better DINO-based structural alignment, enabling cleaner and more controllable weather editing.}
By proactively correcting environmental distortions at the visual input level, we significantly boost the performance of downstream tasks such as object detection and semantic segmentation. We observe that after applying IntrinsicWeather, \revise{the detection and segmentation performance on the ACDC~\cite{sakaridis2021acdc} benchmark increases by 87.1\% ($\mathrm{AP}_{75}$ from 13.15\% to 24.60\%) and 24.5\% ($\mathrm{mIOU}$ from 24.13\% to 30.05\%), respectively}.
\added{We also introduce a synthetic dataset and a real-world dataset containing 38k and 18k images with intrinsic map annotations, covering diverse weather conditions and a wide range of driving scenes to train these two components.}
%\added{Extensive experiments show that our IntrinsicWeather outperforms state-of-the-art methods on several benchmarks.} Moreover, our method demonstrates significant value in downstream tasks for autonomous driving, enhancing the robustness of object detection and semantic segmentation in challenging weather scenarios.
To summarize, our contributions are as follows:
\begin{itemize}
    \item We propose IntrinsicWeather, a method to decompose images into intrinsic maps under various weather conditions, and synthesize them into another lighting or weather condition guided by text prompts. 
    \item We introduce \revise{intrinsic map-aware attention} that provides customized visual detail guidance for generative models, helping our decomposition.
    \item We construct two new datasets called WeatherSynthetic and WeatherReal, containing synthetic and real-world images covering various weather conditions on autonomous driving scenarios, along with their corresponding maps. \yx{The datasets will be released upon acceptance.} 
    % \milos{Can the datasets be released? If so, we should say that.}
\end{itemize}

\section{Related work}
\label{sec:related}

% \paragraph{Generative model.} 
\textbf{\added{Diffusion models.}} 
% A generative model aims to learn the underlying distribution of data and generate new samples that align with the true data. With the rapid development of deep learning, plenty of works have emerged to generate high-quality results using text or image as the condition. 
% \revise{Variational Autoencoder}~\cite{kingma2013auto} leverages variational inference for probabilistic modeling. \revise{Generative Adversarial Network}~\cite{goodfellow2014generative} employs adversarial training to generate high-fidelity outputs, yet suffers from mode collapse and unstable optimization. 
% Diffusion models~\cite{ho2020denoising,rombach2022high,ramesh2022hierarchical,esser2024scaling} have demonstrated the capability to generate high-quality and text-aligned images. 
% Diffusion models mostly follow DDPM~\cite{ho2020denoising} and its enhancements~\cite{song2020denoising,liu2022pseudo,lipman2022flow}, which divide the generation process into forward and reverse processes. In the forward process, the original image is gradually corrupted by adding Gaussian noise. The reverse process employs a neural network, such as UNet~\cite{ronneberger2015unet} or DiT~\cite{peebles2023dit}, to iteratively denoise samples drawn from this Gaussian distribution, ultimately reconstructing the target image. 
Diffusion models have achieved remarkable progress in high-fidelity and text-conditioned image generation~\cite{ho2020denoising,rombach2022high,ramesh2022hierarchical,esser2024scaling}. Modern diffusion frameworks typically adopt a denoising process~\cite{ho2020denoising,song2020denoising,liu2022pseudo,lipman2022flow}, parameterized by UNet- or DiT-based backbones~\cite{ronneberger2015unet,peebles2023dit}.
In our research, we repurpose diffusion models to jointly estimate material, geometry, and lighting from an image while synthesizing new images under specified weather conditions. This demonstrates that the strong priors embedded in pre-trained diffusion models can be effectively leveraged for faithful and physically grounded estimations.

\vspace{-5mm}
\paragraph{Forward and inverse rendering using diffusion.}
The emergence of the diffusion model has catalyzed a novel methodology that leverages generative models to learn the joint probability distribution between images and their corresponding intrinsic maps~\cite{kocsis2024intrinsic,zeng2024rgb,li2024idarb,fu2024geowizard,chen2024uni,liang2025diffusionrenderer,he2025unirelightlearningjointdecomposition,IntrinsicDiffusion}. 
\added{
IID~\cite{kocsis2024intrinsic} focuses on material estimation in indoor scenes, while RGB$\leftrightarrow$X~\cite{zeng2024rgb} extends diffusion to bidirectional mapping between RGB images and maps. Several works further incorporate geometric priors~\cite{fu2024geowizard} or multi-view cues~\cite{li2024idarb}. DiffusionRenderer~\cite{liang2025diffusionrenderer} adapts a video diffusion model to achieve temporally consistent inverse and forward rendering, and UniRelight~\cite{he2025unirelightlearningjointdecomposition} jointly estimates albedo and relighted video frames.
These methods are primarily designed for indoor scenes, small objects, or video relighting, and they struggle to generalize to large-scale outdoor driving scenes with diverse weather conditions. Moreover, none of these works address controllable weather editing or decomposition in intrinsic space under multiple weather conditions, which is the goal of our framework.
}

\vspace{-5mm}
\added{\paragraph{Image and weather editing.}
Most image editing techniques operate purely in the pixel space, using diffusion models to modify color, texture, or local appearance without modeling underlying scene factors~\cite{tumanyan2022plugandplaydiffusionfeaturestextdriven,brooks2023instructpix2pix,wu2025qwenimagetechnicalreport,labs2025flux1kontextflowmatching,hertz2022prompttopromptimageeditingcross,cao2023masactrltuningfreemutualselfattention,parmar2023zeroshotimagetoimagetranslation,kawar2023imagictextbasedrealimage}.
Following this paradigm, existing weather-editing models~\cite{cosne2020using,li2021weatherganmultidomainweather,tasnim2025normalizing} directly translate one weather type to another, and WeatherWeaver~\cite{lin2025controllable} finetunes a video diffusion model for weather removal and synthesis. However, pixel-space editing lacks physical interpretability and cannot guarantee consistent material, geometry, or illumination.
Intrinsic-space manipulation has been explored by IntrinsicEdit~\cite{Lyu_2025}, but it focuses on object-level editing rather than large-scale outdoor scenes. Other approaches edit weather in 3D space~\cite{qian2025weathereditcontrollableweatherediting,li2023climatenerfextremeweathersynthesis}, but they require accurate geometry, which is unavailable for real-world driving scenes.
In contrast, IntrinsicWeather performs controllable weather editing in the intrinsic space. This formulation enables fine-grained and geometry-preserving control that is difficult to achieve in the pixel space.
}

% IID~\cite{kocsis2024intrinsic} firstly trains a latent diffusion model to estimate material, including albedo, roughness, and metallicity. RGB$\leftrightarrow$X~\cite{zeng2024rgb} propose a framework using diffusion models for both FR and IR. DiffusionRenderer~\cite{liang2025diffusionrenderer} finetunes Stable Video Diffusion~\cite{blattmann2023stable} to maintain temporally consistent estimation. Recent UniRelight~\cite{he2025unirelightlearningjointdecomposition} estimates relighted video and albedo jointly.
% However, prevailing methods are predominantly designed for indoor scenes or object-level tasks, whereas autonomous driving environments exhibit dynamic and intricate illumination conditions coupled with expanded scene dimensions. 
% Current approaches struggle to adequately process such vehicular scenarios solely through the inherent generalization capabilities of generative models. 
%The presence of different weather (\eg, rain, snow, fog) in driving environments induces substantial performance degradation in existing systems. To address these challenges, we devise extra priors for diffusion models to empower enhanced decomposition of autonomous driving scenes into constituent intrinsic layers.

\begin{figure*}[htbp]
  \centering
  \includegraphics[width=1.0\textwidth]{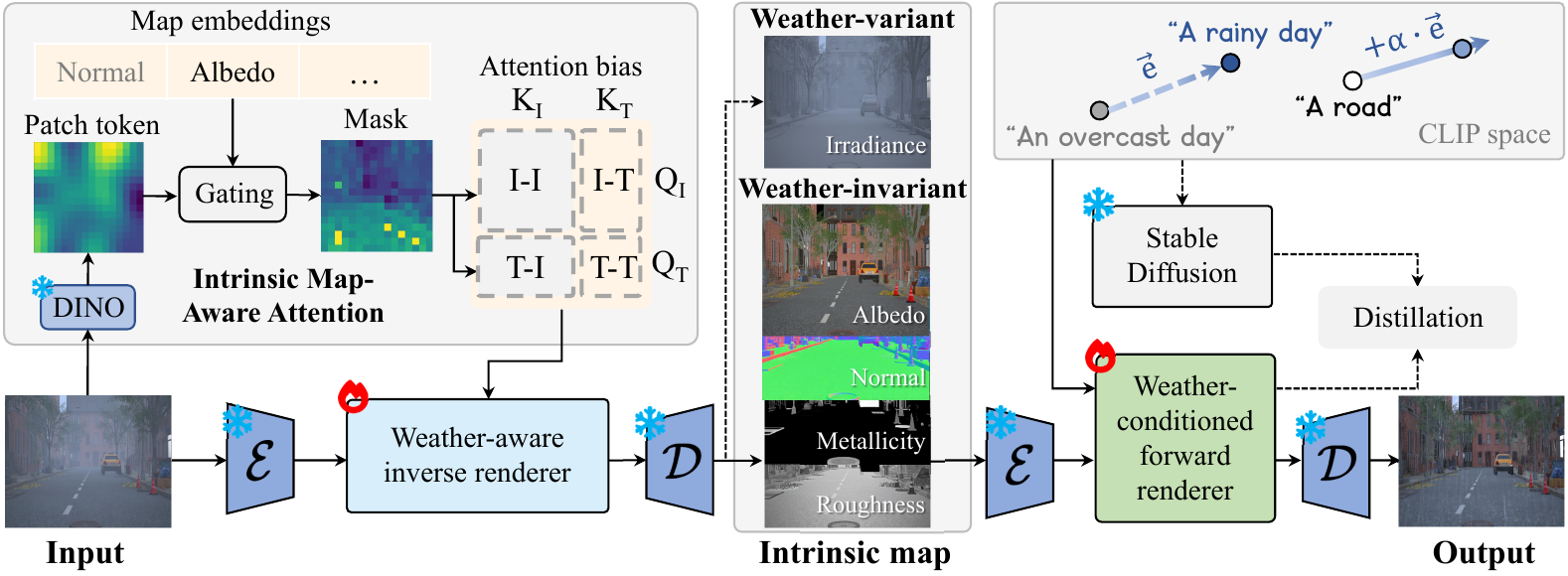}
  \caption{
  \zyx{
  Overview of IntrinsicWeather.
  We propose a diffusion-based framework for controllable weather editing for autonomous driving in intrinsic space. The weather-aware inverse renderer decomposes images into weather-invariant and weather-variant maps, while the weather-conditioned forward renderer re-renders images based on given decomposed maps and text prompts that specify the target condition. 
 For the inverse renderer, we design intrinsic map-aware attention to help the inverse renderer focus on important regions corresponding to target intrinsic maps, where the learned map embeddings filter patch tokens via a gating mechanism. 
For the forward renderer, we design an alpha interpolation in the CLIP semantic space to achieve fine-grained weather control, leveraging the prior in the original Stable Diffusion. By sampling different alpha values, the forward renderer can render natural transitional weather conditions.
  % On the other hand, the FR diffusion renders images based on given decomposed maps and text prompts that specify the target condition, which preserves intrinsic attributes while modifying the weather or materials.
  }
  }
  \label{fig:overview}
\end{figure*}
\section{Method}
\label{sec:method}

\subsection{IntrinsicWeather} \label{subsec:weather-guided diffusion}

\added{We aim to modify weather-related factors \zyx{(\eg, weather particles and accumulations)} of a scene while preserving its underlying geometry and material properties. Pixel-space editing methods inherently entangle weather effects with appearance, making it difficult to maintain structural consistency and natural illumination across different weather conditions. Moreover, weather phenomena are primarily tied to a scene's lighting, rather than intrinsic material attributes. These observations motivate us to move beyond direct pixel manipulation and instead operate in intrinsic space.
}

We propose IntrinsicWeather, a framework designed for controllable weather editing in intrinsic space. The framework includes two components: the weather-aware inverse renderer and the weather-conditioned forward renderer. 
The image is input into the inverse renderer, which performs intrinsic decomposition, disentangling images into weather-invariant material and geometry, as well as weather-variant illumination. Correspondingly, the forward renderer re-renders images based on given intrinsic maps and text prompts that specify the target weather or lighting.

% For the inverse renderer, we design a novel attention mechanism to enforce spatial correspondence between maps and images, which will be introduced in~\cref{subsec: MAA}. 
We repurpose Stable Diffusion 3.5~\cite{sd3.5_medium} to enable the inverse and forward renderers.
To leverage the diffusion prior to achieve fine-grained weather control, we first obtain a weather transitional direction $\boldsymbol{e}$ in the CLIP space:
\begin{equation}
    \boldsymbol{e}=\mathrm{Embed}(\boldsymbol{w}_{1}) - \mathrm{Embed}(\boldsymbol{w}_{2}),
\end{equation}
\yx{where $\boldsymbol{w}_{1}$ denotes the target weather (\eg, rainy), $\boldsymbol{w}_{2}$ denotes the original weather (\eg, overcast), and $\mathrm{Embed}(\cdot)$ is the text encoder.} Then we shift a weather-neutral embedding $\boldsymbol{w}_{base}$ by $\alpha$ steps along this direction:
\begin{equation}
    \boldsymbol{E} = \mathrm{Embed}(\boldsymbol{w}_{base}) + \alpha \cdot \boldsymbol{e}.
\end{equation}
Replacing the original prompt with $\boldsymbol{E}$ can force the model to generate reasonable intermediate results. To preserve the rich priors of the pre-trained model, we align part of the forward renderer's intermediate features with those of the original Stable Diffusion. The detailed implementation of feature distillation is shown in the supplementary material.

\subsection{Intrinsic map-aware attention} \label{subsec: MAA}

\begin{figure}
    \centering
    \includegraphics[width=1\linewidth]{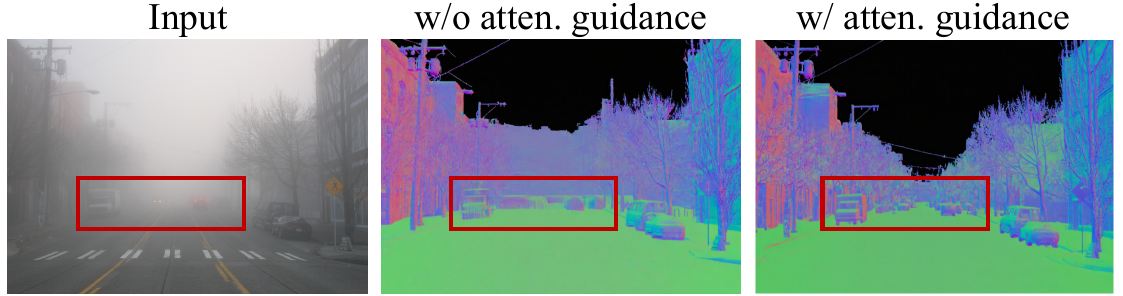}
    \caption{\yx{Attention guidance helps recover distant small objects and fine geometry details.} We present the estimated normal map with and without attention guidance.}
    \label{fig:maa_motivation}
\end{figure}

\begin{figure}[ht]
    \centering
    \includegraphics[width=0.9\linewidth]{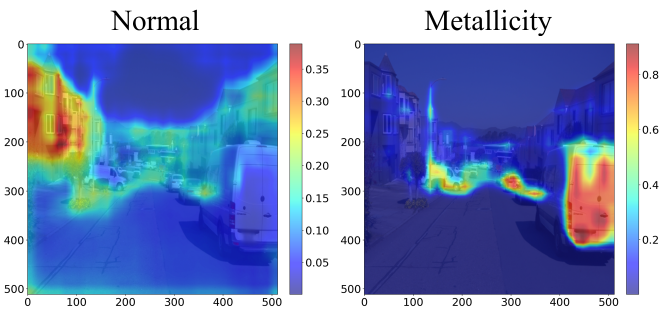}
    \caption{\added{IMAA visualization. Normal estimation primarily concerns the geometry details, especially in regions with sharp variations in surface normals. Metallicity predictions need to selectively attend to metallic objects such as vehicles, poles, and railings. IMAA provides attention guidance for the diffusion model, ensuring spatial correspondence between input and maps.}}
    \label{fig:mask}
\end{figure}

Outdoor and autonomous driving scenes exhibit larger variations in object scale. As shown in~\cref{fig:maa_motivation}, the original Stable Diffusion lacks explicit attention guidance and often performs poorly on distant, small objects and geometrically complex regions. 
% \milos{Sounds like a vague claim, could we point to some specific figure to show this?}
% To this end, we devise attention guidance for DiT with visual priors to capture the structural and semantic information required by the IR.
We observe that different intrinsic maps require attention to distinct regions of an image, as shown in~\cref{fig:mask}.
% \del{For instance, albedo prediction demands a focus on fine-grained object textures, whereas normal estimation primarily concerns the overall orientation of large surfaces such as the ground and walls. Metallicity prediction, in contrast, needs to selectively attend to metallic objects such as vehicles, poles, and railings. 
% Meanwhile, we notice that the patch tokens extracted by DINO \cite{oquab2023dinov2} exhibit strong intra-class consistency, meaning that spatially separated regions corresponding to the same type of material or structure produce highly similar feature representations.}
To leverage these observations, we extend DiT with intrinsic map-aware attention \yx{(IMAA)} to apply attention guidance for DiT.
% \milos{Should it be IMAA?} 

% First, we explore how to filter important regions according to intrinsic maps. Then, we construct attention bias from these regions.
We identify important regions based on the intrinsic maps and use them to construct an attention bias.
Specifically, we first employ DINOv2 \cite{oquab2023dinov2} to extract a set of patch tokens $\boldsymbol{p}$. For each intrinsic map, we define a learnable embedding $\boldsymbol{d} \in \mathbb{R} ^ {D_{\text{model}}}$ that captures its inherent characteristics. A gating mechanism is applied to selectively filter patch tokens based on the current intrinsic map. Formally, we compute a map-aware mask:
\begin{equation}
    \boldsymbol{m} = \mathrm{gating}(\boldsymbol{p}, \boldsymbol{d}) = \mathrm{MLP}\big([f_{p}(\boldsymbol{p}),f_{\boldsymbol{d}}(\boldsymbol{d})]\big),
    \label{eq:map-aware attn}
\end{equation}
\yx{where $f_p(\cdot)$ and $f_d(\cdot)$ are linear projections of patch tokens $\boldsymbol{p}$ and map embedding $\boldsymbol{d}$, $[\cdot]$ indicates the concatenated input to the MLP. This gating mechanism highlights image regions most relevant to the target map.
% Intuitively, this gating mechanism produces a mask that emphasizes regions in the image that are most relevant to the target intrinsic map, effectively “gating” the attention of the model toward important regions.
}
% where $\boldsymbol{m}$ is the map-aware mask calculated by the gating mechanism, which applies feature modulation between map embedding $\boldsymbol{d}$ and patch tokens $\boldsymbol{p}$. 
% \milos{There are too many symbols left unexplained here: $f_p$, $f_d$, $||$, square brackets, gating...}

We construct a joint attention bias $\boldsymbol{M}$ using $\boldsymbol{m}$ to guide the diffusion model, effectively enhancing the attention logits in DiT. 
We apply the map-aware mask to the text–image and image–image parts of the joint attention matrix: the former enforces textual guidance on important regions, while the latter strengthens feature aggregation within the image space. 
% Formally, the bias $\boldsymbol{M}$ is given by:
% \begin{equation}
%     \boldsymbol{M}_{i, j} = 
%     \begin{cases}
%         \boldsymbol{m}_{i}, & \text{if } i \leq \textrm{len}(\boldsymbol{K}_{I}) \\
%         0, & \text{else}
%     \end{cases}
% \end{equation}
% where $i, j$ is the index of token, $\boldsymbol{K}_{I}$ is the image token. 
\yx{
Formally, the bias $\boldsymbol{M}$ is defined as:
\begin{equation}
    \boldsymbol{M}_{i,j} =
    \begin{cases}
        \boldsymbol{m}_{i}, & \text{if } i \text{ indexes an image token in } \boldsymbol{K}_{I}, \\
        0, & \text{otherwise}.
    \end{cases}
\end{equation}
where $i$ and $j$ denote token indices, and $\boldsymbol{K}_{I}$ is the set of image tokens.
}

Then the bias is applied to the DiT: 
\begin{equation}
    \mathrm{Attn}(\boldsymbol{Q}, \boldsymbol{K}, \boldsymbol{V}) = \mathrm{Softmax}\big(\frac{\boldsymbol{Q}\boldsymbol{K}^T}{\sqrt{d_k}} + \boldsymbol{M}\big)\boldsymbol{V},
\end{equation}
where $\boldsymbol{Q}, \boldsymbol{K}, \boldsymbol{V}$ are combination of image and text token.

\revise{Moreover, we devise a heuristic-guided progressive training strategy for IMAA to stabilize learning and ensure that IMAA provides meaningful guidance in the early stages. For example, we use the gradient operator to extract illuminated regions and shadow boundaries for the irradiance map. The detailed description and ablation studies are shown in the supplementary material.}

\subsection{Dataset construction} \label{subsec: Dataset}
\begin{figure}
    \centering
    \includegraphics[width=1.0\linewidth]{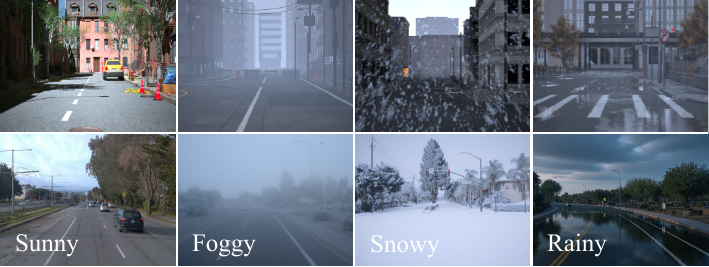}
    \caption{ Example of our WeatherSynthetic (the first row) and WeatherReal (the second row).}
    \label{fig:dataset}
\end{figure}

% To train and evaluate IntrinsicWeather, we require large-scale datasets that provide both diverse weather conditions and intrinsic ground truth.
Existing datasets of images with corresponding intrinsic maps suffer from the absence of outdoor environments (OpenRooms~\cite{li2020inverse}, Hypersim~\cite{roberts2021hypersim}, InteriorVerse~\cite{zhu2022learning}) or insufficient weather diversity (MatrixCity~\cite{li2023matrixcity}), and thus are inappropriate for large-scale outdoor and autonomous driving scenes. 
\revise{While MatrixCity provides continuous variations of fog density and illumination, it mainly targets relighting and reconstruction tasks. In contrast, our work requires diverse weather and lighting conditions rather than smooth transitions of the same type.}
Moreover, MatrixCity suffers from legal licensing issues. To fill the gap in large-scale weather-diverse autonomous driving datasets with paired images and intrinsic maps, we propose WeatherSynthetic and WeatherReal. Sample images from our datasets are shown in~\cref{fig:dataset}.
% Considering the current lack of a dataset dedicated to large-scale inverse rendering tasks in autonomous driving scenarios, where OpenRooms~\cite{li2020inverse}, Hypersim~\cite{roberts2021hypersim}, and InteriorVerse~\cite{zhu2022learning} are limited to indoor environments, and MatrixCity~\cite{li2023matrixcity} only covers urban scenes with insufficient weather diversity, we introduce WeatherSynthetic and WeatherReal. They include various weather types and rich corresponding intrinsic maps, and the overview of our datasets is shown in \hyperref[tab:dataset]{Table.~\ref{tab:dataset}}.

\vspace{-2mm}
\paragraph{WeatherSynthetic}
is a large-scale synthetic dataset encompassing a wide range of scene and weather types: 
\begin{itemize}
  \item \textbf{Weather:} sunny, overcast, rainy, thunderstorm, snowy, foggy, sandstorm.
  \item \textbf{Time of day:} early morning, morning, noon, afternoon. 
  \item \textbf{Environment:} urban, suburban, highway, parking
\end{itemize}
We use Unreal Engine 5 to render all images and intrinsic maps. We purchased 3D assets that are cleared for generative model use in Fab. The rendering pipeline uses the movie render queue and multi-sample anti-aliasing, producing high-quality rendering results. The UltraDynamicSky and UltraDynamicWeather are applied to modify the weather and daytime. 
Note that all images are in linear space without tone mapping. 
%\del{This way, we get 35K images at the cost of 24h.} 
\revise{In total, rendering the 38K images and maps took about 24 hours on our setup.} 
% Some typical scenes are shown in ~\cref{fig:dataset}.}
% We use UltraDynamicSky and UltraDynamicWeather to simulate different daytime and weather in UE5. High-quality images are rendered using the Movie Render Queue, along with corresponding physical and material property annotations. To achieve high quality of our rendered image and intrinsic maps, we set the anti-aliasing method as Multi-Sample Anti-Aliasing. All images in our dataset are in linear space without tone mapping. Finally, we get 35 thousand images spending nearly 24 hours, and some typical scenes are shown in \hyperref[fig:dataset]{Fig.~\ref{fig:dataset}}. 
%Although there is a distributional difference between indoor datasets and our task scenario, we believe they can still help the model learn some image decomposition capabilities. Therefore, in addition to the WeatherDrive, we also used InteriorVerse~\cite{zhu2022learning} and Hypersim~\cite{roberts2021hypersim}.

\vspace{-5mm}
\paragraph{WeatherReal}
is a real-world dataset on autonomous driving scenes with various weather conditions. We use our inverse renderer to generate intrinsic maps of open-source datasets such as Waymo \cite{sun2020scalability} and Kitti~\cite{geiger2013vision}. 
% Since these datasets were collected under sunny conditions, the model trained on WeatherSynthetic is adequate for generating pseudo ground truth. 
We use a multimodal model to remove challenging scenarios (e.g., rainy nights) to ensure the generation of high-quality pseudo-labels, and check the quality manually. The samples are shown in the supplementary materials.
Moreover, we employ a pre-trained image editing model (\ie, Flux-Kontext~\cite{labs2025flux1kontextflowmatching}) to alter the weather types. 
Our WeatherReal is motivated by the observation that after training our model merely on synthetic data, the inverse and forward renderers lack sufficient generalization capability on real-world samples. \added{Note that WeatherReal is only used to finetune models and is not used to evaluate results.}

\begin{figure*}
    \centering
    \includegraphics[width=1\textwidth]{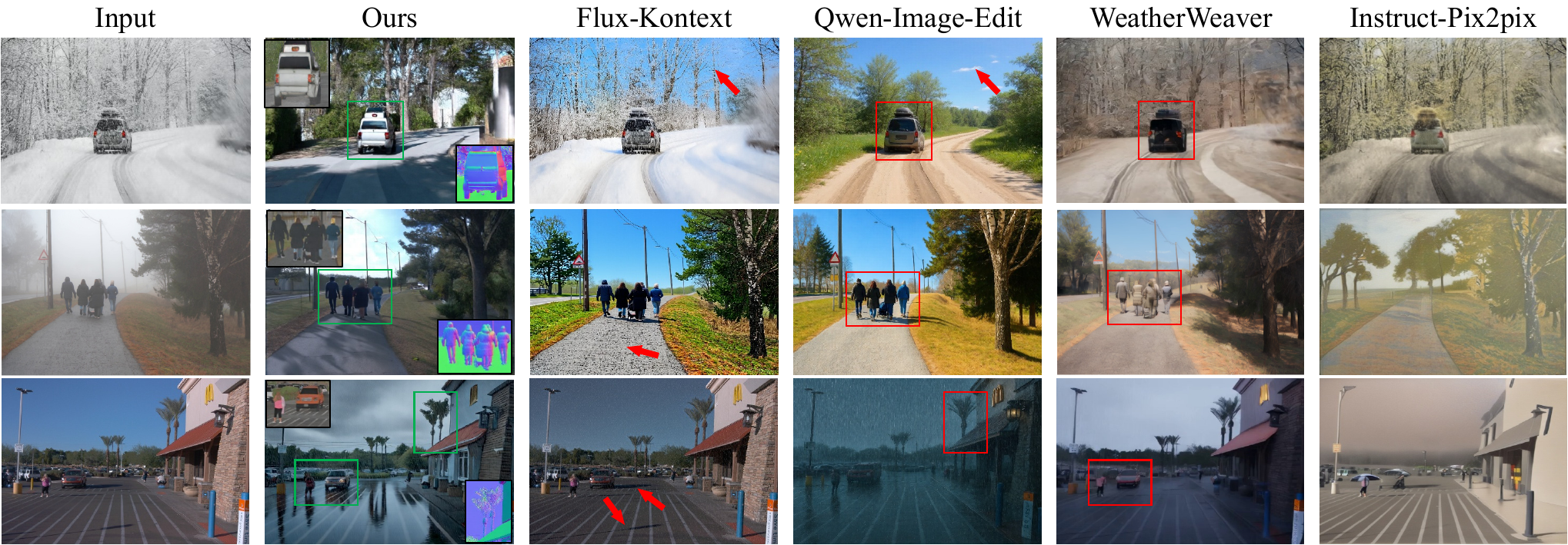}
    \caption{
    \added{
    Comparison with pixel-space editing methods. 
    We use the prompt ``A sunny/rainy day'' for our model. For pixel-space editing methods, we use their recommended instruction format, \ie, ``Turn weather into a sunny/rainy day''.
    Pixel-space methods struggle to preserve scene geometry and materials, often introducing hallucinated objects, distorted structures, or unnatural lighting.
    In contrast, our intrinsic-space editing preserves geometry and appearance while modifying only weather-related components.
    For clarity, we highlight corresponding artifacts with red boxes and arrows. We also show intrinsic maps (albedo + normal) for our method, demonstrating explicit disentanglement of material, geometry, and illumination.
    % Our method removes all the snow cover and snowflakes while preserving the car's color and distant trees. Other image editing methods struggle to remove all the particles and accumulations and recover material and geometry. For the rainy scene, our method generates natural reflections and raindrops. Flux-Kontext fails to edit the ground, and Qwen-Image-Edit generates brushstroke-like textures.
    }
    }
    \label{fig:compare_with_editing}
\end{figure*}

\section{Experimental results}

% Our IntrinsicWeather first leverages the inverse renderer to obtain intrinsic maps of the input, and then uses the forward renderer to re-render the image based on the maps and text descriptions.
% In this section, we first evaluate the inverse and forward rendering on both synthetic and real-world datasets, followed by comparisons with pixel-space editing and weather restoration methods. We then conduct ablation studies on MAA and datasets, and conclude with a discussion of the limitations of our approach.
In this section, we first compare our method with pixel-space editing and weather restoration methods. Then we evaluate the inverse and forward rendering. We then conduct ablation studies on IMAA and datasets, and conclude with a discussion of the limitations of our approach.

Following WeatherWeaver~\cite{lin2025controllable}, we use PickScore~\cite{kirstain2023pickapicopendatasetuser}, CLIP image-text consistency (denoted as CLIP-S), and DINO structure similarity (denoted as DINO-S) to evaluate editing results.
Following previous works \cite{zeng2024rgb,li2024idarb}, we report Peak Signal-to-Noise Ratio (PSNR), Structural Similarity Index Measure (SSIM), Mean Angular Error (MAE), and Learned Perceptual Image Patch Similarity (LPIPS) for inverse rendering. 
We compare our performance with pixel-space editing methods (Flux-Kontext~\cite{labs2025flux1kontextflowmatching}, Qwen-Image-Edit~\cite{wu2025qwenimagetechnicalreport}, Instruct-Pix2Pix~\cite{brooks2023instructpix2pix}, WeatherWeaver~\cite{lin2025controllable}) and weather restoration methods (AWRaCL~\cite{rajagopalan2025awracle}, Histoformer~\cite{sun2024restoringimagesadverseweather}).
We also compare our inverse and forward rendering results with RGB$\leftrightarrow$X~\cite{zeng2024rgb}, IID~\cite{kocsis2024intrinsic}, Geowizard~\cite{fu2024geowizard}, IDArb~\cite{li2024idarb} and DiffusionRenderer~\cite{liang2025diffusionrenderer}. We evaluate different methods on WeatherSynthetic, Waymo~\cite{sun2020scalability}, TransWeather~\cite{valanarasu2021transweather}, ACDC~\cite{sakaridis2021acdc}, and additional Internet images covering diverse weather conditions.

\subsection{Comparison with pixel-space editing methods} 
 We show quantitative comparisons in \cref{tab:forward_fid}. Our method achieves the highest CLIP-S, indicating that it produces the most text-aligned and plausible weather editing results. In terms of DINO-S, we rank second only to Flux-Kontext~\cite{labs2025flux1kontextflowmatching}, which, however, fails to remove or synthesize weather effects effectively. Although Qwen-Image-Edit~\cite{wu2025qwenimagetechnicalreport} achieves a slightly higher PickScore, it often introduces inconsistent textures and geometry. PickScore is suitable for measuring user preference, but it does not measure physical consistency and editing plausibility.
 
 A qualitative comparison is shown in~\cref{fig:compare_with_editing}. 
 In the first row, our method removes all the snowflakes and snow accumulation on the trees. Flux-Kontext fails to remove them, while Qwen-Image-Edit and WeatherWeaver mistakenly change the geometry of the scene and the car's color. For the second row, our method removes the dense mist, recovering the color and pose of pedestrians. Flux-Kontext generates noisy textures while Qwen-Image-Edit and WeatherWeaver change the count and pose of pedestrians. In the last row, we transform the weather into a rainy day, generating natural reflection and lighting. The other methods produce unnatural lighting while struggling to preserve geometry and material. Flux-Kontext adds some rain streaks on the original image, making the sunny-day shadows on the ground look noticeably out of place. Instruct-Pix2Pix struggles to manipulate weather effects and instead performs incorrect operations resembling style transfer. 
 
 We also present the albedo and normal map obtained from our inverse renderer. Our re-rendered images align well with these maps. 
 Weather editing in intrinsic space allows our model to completely remove weather-related artifacts, including both airborne particles and surface accumulations, while preserving geometric and material consistency. 
 Furthermore, the disentangled material and geometry obtained from inverse rendering facilitate realistic illumination and shadow generation during re-rendering. 
 More comparisons are shown in the supplementary.

\begin{figure}
    \centering
    \includegraphics[width=1\linewidth]{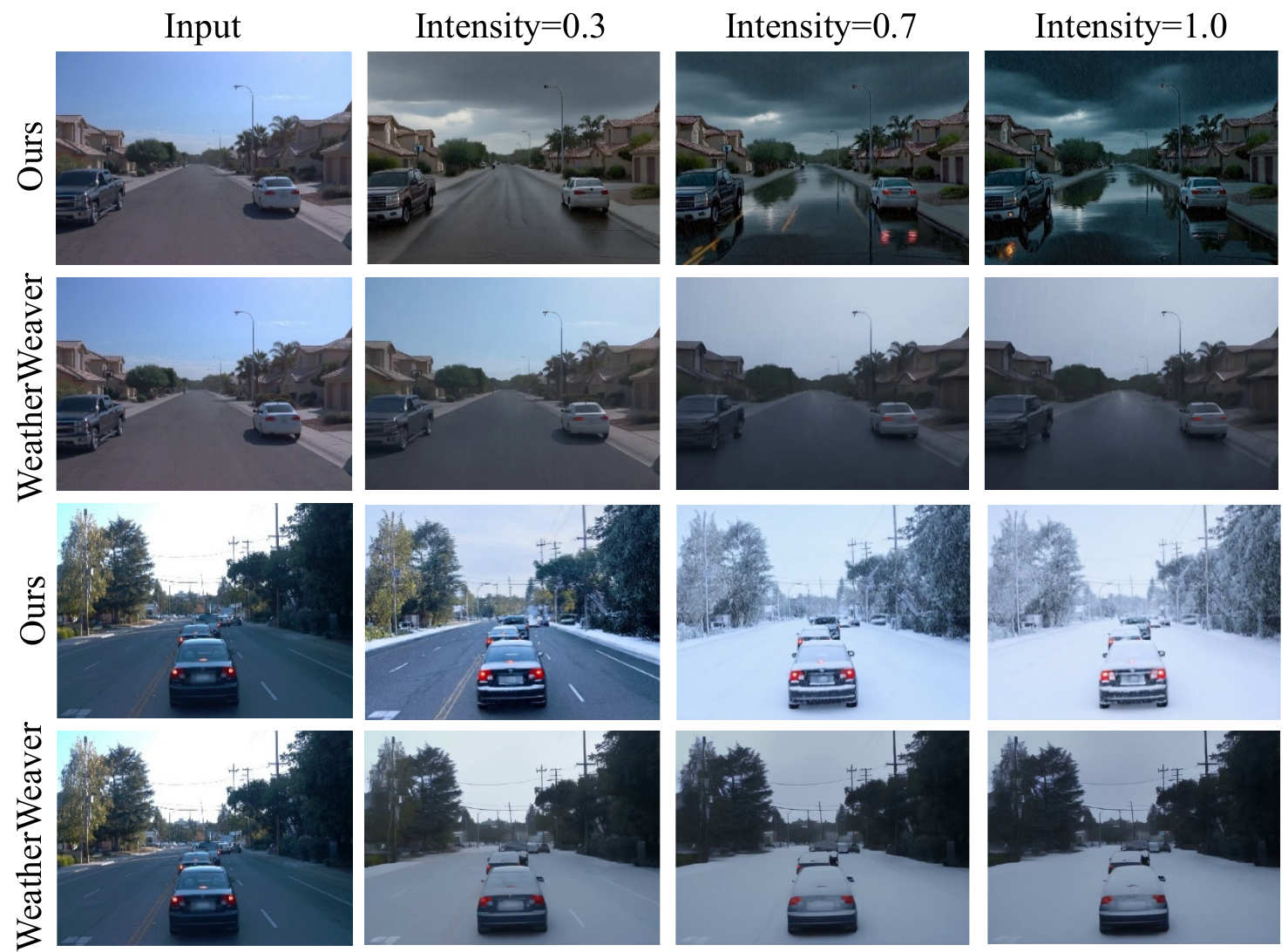}
    \caption{Comparison of fine-grained weather control. Our editing results show natural transitions. WeatherWeaver shows lower realism and looks closer to a blending effect.}
    \label{fig:supp_finegrain}
\end{figure}

 We further compare our method with WeatherWeaver on fine-grained weather control in~\cref{fig:supp_finegrain}. Our editing results show natural transitions: under light to heavy rain, the road surface gradually becomes wetter; under different levels of snowfall, snow first appears along the roadside and on branches, and eventually accumulates to cover the entire road. In contrast, WeatherWeaver lacks this sense of realism and instead looks more like applying filters of increasing intensity to the original image. 
 \vspace{-5mm}
\zyx{
\paragraph{User study.} We conducted a user study to evaluate the consistency and realism of different editing methods. A total of 61 participants were asked to vote on 8 cases covering weather removal and weather synthesis. The average preference for our results is 81.67\%, showing that our method is preferred by users. The detailed setup and results are provided in the supplementary material.
}

\begin{table}[t]
    \centering
    \caption{\added{Comparison with rendering-based methods and pixel-space editing methods. 
    % Our method gets the highest CLIP-S and ranks second best for structure consistency (DINO-S), achieving the best balance between editing plausibility and structural consistency. Although large-scale image editing models such as Flux-Kontext and Qwen-Image-Edit achieve slightly higher scores, they often introduce inconsistent geometry and textures (see~\cref{fig:compare_with_editing}).
    }
    }
    \resizebox{\linewidth}{!}{
    \begin{tabular}{lcccccc}
        \toprule
        Method & \multicolumn{4}{c} {PickScore $\uparrow$~\cite{kirstain2023pickapicopendatasetuser}} & CLIP-S $\uparrow$ & DINO-S $\uparrow$\\
        \cmidrule(lr){2-5} 
             & Sunny  & Snowy & Foggy & Rainy  &\\
        \midrule
         \multicolumn{7}{l}{\textit{Rendering-based methods}} \\
         Ours  & 20.59 & \subsota{22.32} & \subsota{21.34} & \subsota{20.76} & \sota{27.66} & \subsota{73.63} \\
         RGB$\leftrightarrow$X  & 20.40 & 19.92 & 19.71 & 20.29 & 19.00 & 55.87\\
         DiffusionRenderer & 20.24 & -- & -- & -- & -- & 43.12\\
         \midrule
         \multicolumn{7}{l}{\textit{Pixel-space editing methods}} \\
         Flux-Kontext & \subsota{20.72} & 22.25 & 20.99 & 19.46 & 24.46 & \sota{85.50} \\
         Qwen-Image-Edit & \sota{20.77} & \sota{22.43} & \sota{21.82} & \sota{21.56} & \subsota{27.14} & 53.70\\
         WeatherWeaver & 20.13 & 21.41 & 20.93 & 20.25 & 26.78 & 67.01 \\
        Instruct-Pix2Pix & 20.27 & 21.39 & 20.98 & 19.83 & 23.79 & 22.41 \\
        \bottomrule
    \end{tabular}
    }
    \label{tab:forward_fid}
\end{table}

 \begin{figure*}
    \centering
    \includegraphics[width=1\textwidth]{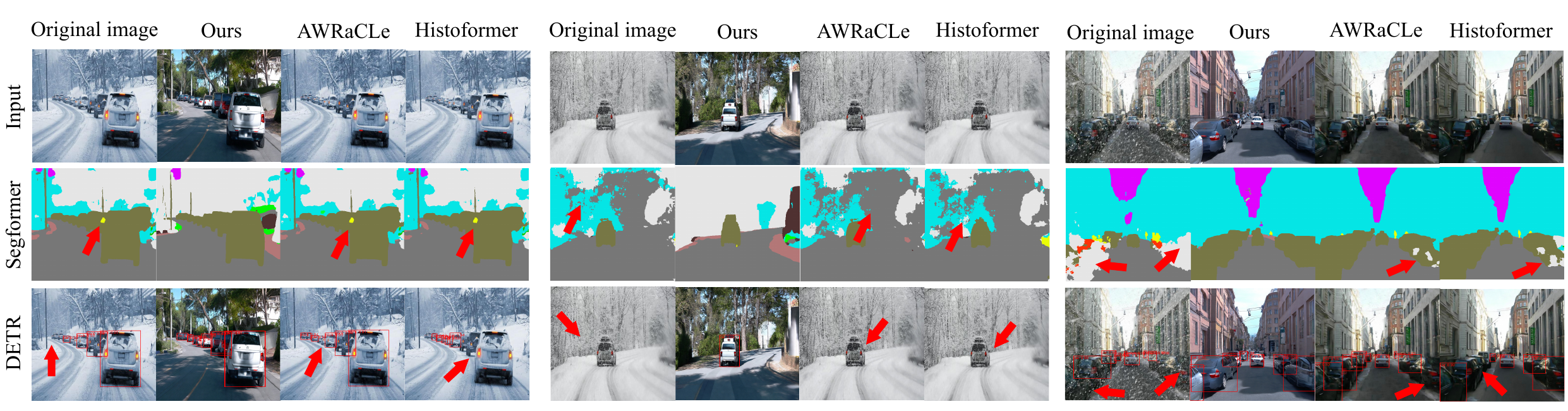}
    \vspace{-5mm}
    \caption{
    Validation of the enhancement of detection and segmentation. Our re-rendered images (prompt: ``The image depicts a bright sunny day.'') not only remove airborne particles (\eg, snowflakes) but also restore material and lighting conditions (\eg, removing surface snow), leading to more accurate segmentation and detection results. In contrast, AWRaCLe and Histoformer primarily remove particles in the air but fail to correct material or lighting degradations. Red arrows note the wrong estimations.
    % Forward rendering on real data with different weather conditions. We show the original images and re-rendered images in the first line. The following lines show segmentation and detection results. SegFormer~\cite{xie2021segformer} and DETR~\cite{carion2020detr} both fail to give reasonable estimates under bad weather conditions, while their performance improves on the re-rendered images generated by our IntrinsicWeather. Moreover, we present a comparison with the state-of-the-art weather restoration method (AWRaCLE, AAAI 2025)~\cite{rajagopalan2025awracle}. 
    % The advantage of our method is its ability to effectively remove various degradations in adverse weather images, such as airborne particles (like snowflakes) and snow on the ground.
    }
    \label{fig:real_forward}
\end{figure*}

\subsection{Comparison with weather restoration methods}
We present qualitative comparisons with AWRaCLe~\cite{rajagopalan2025awracle} and Histoformer~\cite{sun2024restoringimagesadverseweather} in \cref{fig:real_forward}. The advantage of our method is its ability to effectively remove various degradations in adverse weather images, such as airborne particles (like snowflakes), snow on the ground, and overall illumination. The weather restoration methods only remove particles while failing to change surface material or lighting conditions. 

\added{To further verify the improvement brought by different models to downstream applications, we choose object detection and semantic segmentation as tasks, evaluating performance improvement before and after weather editing. Specifically, we chose DETR~\cite{carion2020detr} and Segformer~\cite{xie2021segformer} as base models.} \revise{When applying object detection and semantic segmentation models to our re-rendered images, both tasks achieve more accurate and consistent results, as shown in the lower rows of \cref{fig:real_forward}.} \added{Following this way, we apply weather editing and evaluate on the validation set of the ACDC benchmark~\cite{sakaridis2021acdc}, and the results are shown in~\cref{tab:application}.}

\begin{table}[t]
\centering
\caption{Object detection and semantic segmentation results on the ACDC validation set before and after applying IntrinsicWeather.}
\label{tab:application}
\resizebox{0.85\linewidth}{!}{
\begin{tabular}{lcccc}
\toprule
 & $\text{AP}_{0.5}$ & $\text{AP}_{0.75}$ & $\text{mAP}_{[0.5:0.95]}$ & mIOU\\
\midrule
DETR  & 56.56 & 13.15 & 47.00 & --\\
DETR + ours   & 61.32 & 24.60 & 54.87 & --\\
\midrule
Segformer  & -- & -- & -- & 24.13 \\
Segformer + ours   & -- & -- & -- & 30.05 \\
\midrule
Absolute Gain & +4.76 & +11.45 & +7.87 & +5.92 \\
\bottomrule
\end{tabular}
}
\end{table}

\begin{table*}
    \centering
    \caption{Quantitative evaluations of our method against existing methods in terms of decomposition quality on the WeatherSynthetic test set. Considering that IID and RGB$\leftrightarrow$X were only trained on indoor datasets, we finetune them on our WeatherSynthetic and show the results before and after finetuning. We highlight the best results in \sota{red} and the second-best ones in \subsota{orange}.}
    \resizebox{\textwidth}{!}{
    \begin{tabular}{lcccccccccccc}
        \toprule
         Method                
         & \multicolumn{3}{c}{Albedo} 
         & \multicolumn{3}{c}{Normal} 
         & \multicolumn{2}{c}{Roughness} 
         & \multicolumn{2}{c}{Metallicity} 
         & \multicolumn{2}{c}{Irradiance} \\
        \cmidrule(lr){2-4} \cmidrule(lr){5-7} \cmidrule(lr){8-9} \cmidrule(lr){10-11} \cmidrule(lr){12-13}
         & PSNR $\uparrow$  & SSIM $\uparrow$ & LPIPS $\downarrow$ 
         & PSNR $\uparrow$  & SSIM $\uparrow$ & MAE $\downarrow$ 
         & PSNR $\uparrow$  & LPIPS $\downarrow$ 
         & PSNR $\uparrow$  & LPIPS $\downarrow$ 
         & PSNR $\uparrow$  & LPIPS $\downarrow$ \\
        \midrule
         IID                   & 7.80 & 0.26 & 0.63 & -- & -- & -- & 10.30 & 0.55 & 12.37 & 0.64 & -- & -- \\
         IID (w/ finetune)     & 11.55 & 0.53 & 0.40 & -- & -- & -- & 12.34 & 0.43 & 12.22 & 0.55 & -- & -- \\
         RGB$\leftrightarrow$X & 9.66 & 0.44 & 0.47 & 11.90 & 0.41 & 15.51 & 13.62 & 0.55 & -- & -- & 16.24 & 0.58 \\
         RGB$\leftrightarrow$X (w/ finetune) & 11.35 & 0.59 & 0.37 & 16.14 & 0.49 & 7.05 & 13.65 & 0.57 & 11.96 & 0.66 & 16.38 & 0.69 \\
         GeoWizard             & -- & -- & -- & 16.24 & 0.54 & 12.47 & -- & -- & -- & -- & -- & -- \\
         IDArb                 & 6.40 & 0.48 & 0.65 & 10.77 & 0.43 & 22.42 & 10.70 & 0.62 & 14.66 & 0.62 & -- & -- \\
         DiffusionRenderer     & 11.91 & 0.64& \sota{0.34} & 16.43 & \subsota{0.70} & 28.68 & 11.31 & 0.42 & 10.05 & 0.43 & -- & -- \\
         \midrule
         Ours                 & \sota{27.99} & \sota{0.86} & \subsota{0.35} & \sota{25.06} & \sota{0.84} & \sota{4.24} & \sota{25.81} & \sota{0.23} & \sota{29.29} & \sota{0.04} & \sota{29.66} & \sota{0.22} \\
         Ours (w/o IMAA)       & \subsota{26.78} & \subsota{0.84} & 0.43 & \subsota{23.63} & 0.79 & \subsota{6.33} & \subsota{24.60} & \subsota{0.25} & \subsota{28.16} & \subsota{0.05} & \subsota{26.99} & \subsota{0.32} \\
        \bottomrule
    \end{tabular}
    }
    \label{tab:evaluation_driving}
\end{table*}

\subsection{Evaluation for components}
\subsubsection{Inverse rendering}
In this part, we first conduct quantitative evaluations on WeatherSynthetic, then we evaluate on real-world TransWeather~\cite{valanarasu2021transweather} datasets.
% \paragraph{Synthetic data.}
We show the comparison between our method and existing methods on the test set of our WeatherSynthetic in ~\cref{tab:evaluation_driving}. Our method outperforms existing approaches across all evaluation metrics. We fine-tune IID~\cite{kocsis2024intrinsic} and RGB$\leftrightarrow$X~\cite{zeng2024rgb} with the same training steps on our WeatherSynthetic. Their performance improves, but they fail to provide high-quality estimation. An overall qualitative comparison is shown in the supplementary.
% Qualitative results are shown in ~\cref{synthetic_inverse}. We compare our method with others under different weather conditions.
% Across both sunny and adverse weather conditions, other methods struggle, especially under fog, snow, and rain, while ours consistently produces clean and faithful predictions.
% On a sunny day (row 1), both ours and DiffusionRenderer produce reasonable estimations, whereas the other methods fail.
% Under adverse weather conditions (rows 2–5), the performance of all other methods deteriorates significantly due to the presence of airborne particles.
% In contrast, our method effectively removes the influence of fog, snowflakes, and rain, yielding clean and faithful predictions. 

\begin{figure}
    \centering
    \includegraphics[width=1\linewidth]{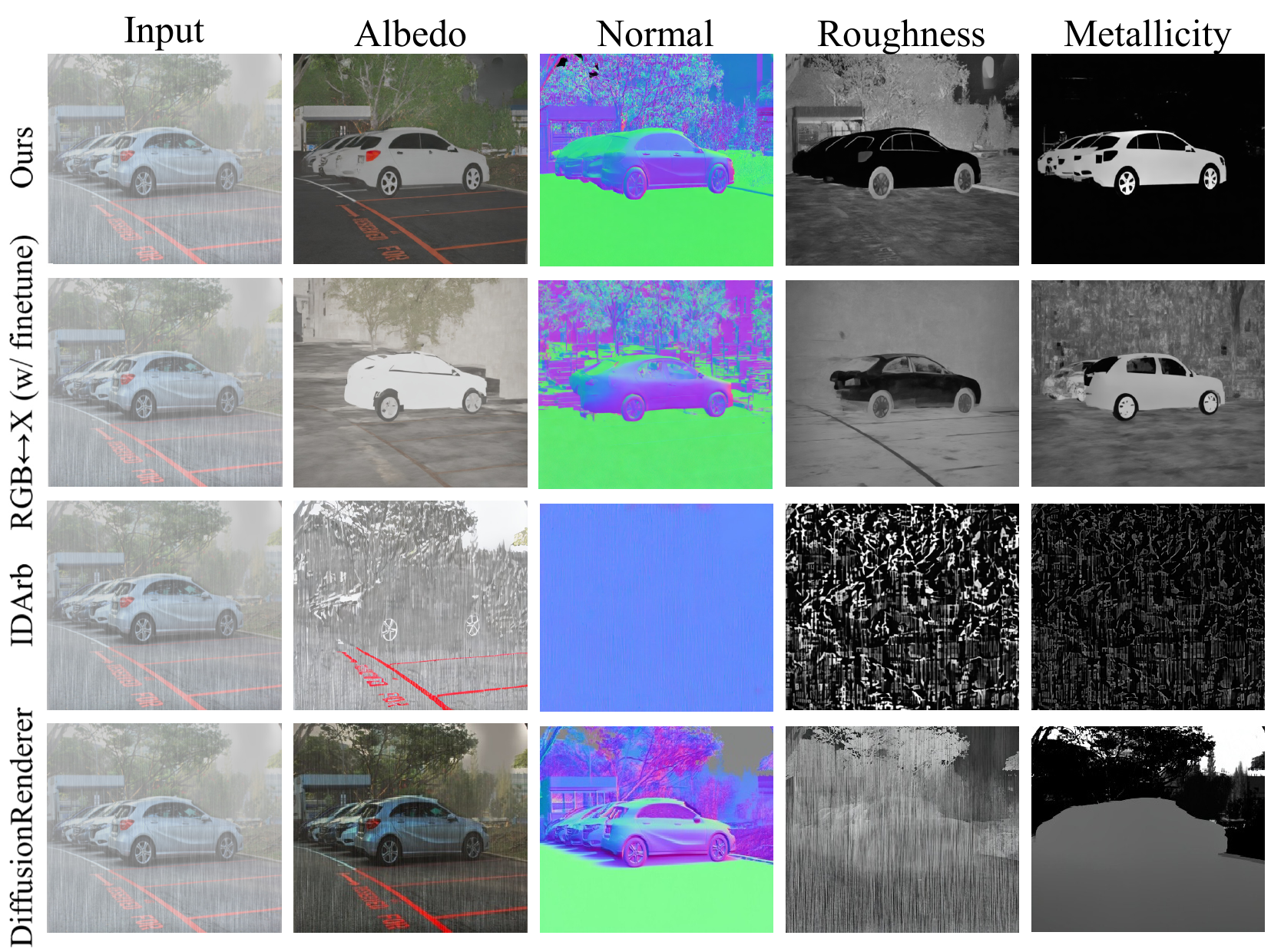}
    \caption{Qualitative comparison of inverse rendering on real-world data. All other methods are affected by rain, but ours removes the disturbance and generates a reasonable estimation. Other map comparisons are shown in the supplementary material.}
    \label{fig:real_inverse}
\end{figure}

We show a comparison of real images of heavy rain in \cref{fig:real_inverse}. All other methods fail to give faithful estimations, while our IntrinsicWeather provides reasonable estimations. 
% We show more results in the supplementary material.
We further validate the consistency of our inverse renderer across weather conditions in the supplementary materials. For each scene, we run the inverse renderer on images captured under different weather types and compute the PSNR between the recovered intrinsic maps and those obtained under sunny conditions.

\begin{figure}
    \centering
    \includegraphics[width=1\linewidth]{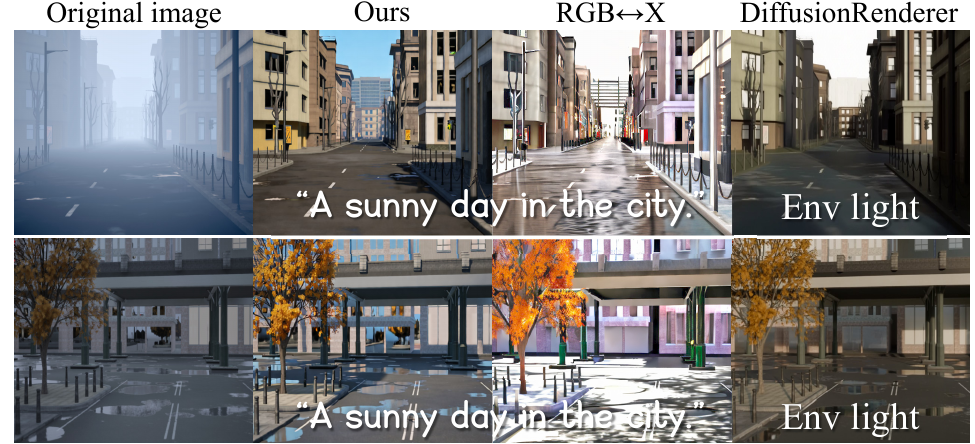}
    \caption{Comparison of forward rendering results. \revise{We use the inverse renderer to obtain intrinsic maps of the original images, and then use the forward renderer to re-render images.} 
    % Our rendered images recover the original details and present natural illumination. RGB$\leftrightarrow$X and DiffusionRenderer produce overexposed or underexposed lighting and miss some details.
}
    \label{fig:synthetic_forward}
\end{figure}

\subsubsection{Forward rendering}
% \paragraph{Synthetic data.}
A comparison of forward rendering between IntrinsicWeather, RGB$\leftrightarrow$X, and DiffusionRenderer~\cite{liang2025diffusionrenderer} is shown in \cref{fig:synthetic_forward}. We use the prompt ``A sunny day in the city.'' for ours and RGB$\leftrightarrow$X, and provide an environment lighting for DiffusionRenderer. Our IntrinsicWeather produces images that better align with the text description than RGB$\leftrightarrow$X, and avoid abnormal textures and illuminations. DiffusionRenderer fails to recover all details, such as road signs and distant buildings.
As shown in \cref{tab:forward_fid}, our method achieves the highest metrics in rendering-based methods on all weather conditions. More results are shown in the supplementary.

\begin{figure}
    \centering
    \includegraphics[width=1\linewidth]{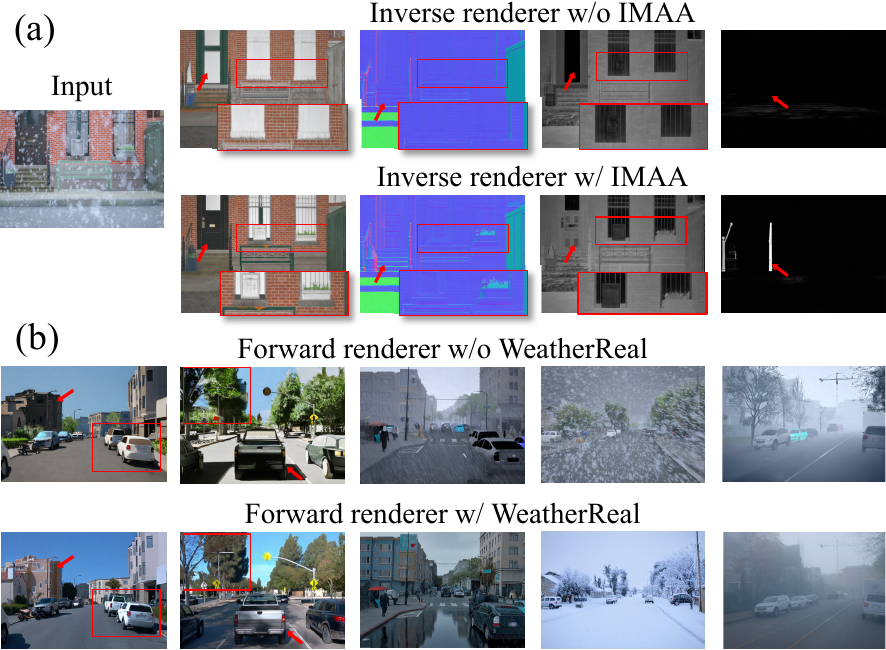}
    \caption{Ablation study. The inverse renderer with IMAA can focus on the details of the original image and recover detailed information. The forward renderer with WeatherReal can generate more realistic lighting, objects, and weather effects.}
    \label{fig:ablation}
\end{figure}

\subsection{Ablation study}
\paragraph{Effect of IMAA.} 
% As highlighted in ~\Cref{sec:method}, a key observation of our method is that different intrinsic maps should pay attention to different regions of the image. 
% Our MAA provides attention guidance for diffusion to help improve the quality of inverse rendering. 
We train an inverse renderer without IMAA for the same steps. As shown in \cref{tab:evaluation_driving}, the model without IMAA behaves poorly than our full model. We show a qualitative result in part (a) of~\cref{fig:ablation}. With the attention guidance related to the map provided by IMAA, the model produced more refined geometry and material predictions and successfully identified the metallic handrail, assigning it a reasonable level of metallicity.

\vspace{-5mm}
\paragraph{Effect of WeatherReal.}
We explore the effect of WeatherReal. We train the forward renderer with only the synthetic dataset and with the WeatherReal dataset. The qualitative results are shown in part (b) of~\cref{fig:ablation}. We use the same intrinsic maps as input and evaluate each model. After training solely on the synthetic dataset, the forward renderer fails to reach high realism, resulting in unrealistic lighting and objects. After introducing WeatherReal, the model learn the distribution of the real-world data and then generates high-quality rendered images. We explore more ablation of datasets in the supplementary material.

\vspace{-5mm}
\paragraph{Effect of intrinsic representation.}
We explore the necessity of intrinsic representation. We replace the intrinsic with a clean-weather image like WeatherWeaver~\cite{lin2025controllable}, and we find that the model struggles to maintain scene details without physical inductive bias. The detailed analysis and results are available in the supplementary.

\subsection{Limitations}
\added{Our framework is designed for single-image weather editing in the intrinsic space.
Temporal modeling introduces additional factors, such as object motion and occlusion changes, that are orthogonal to our core contribution and require a video prior.
Therefore, the current framework does not guarantee temporal consistency in video sequences. Extending IntrinsicWeather to videos would likely be possible by building on a video diffusion model, similar to DiffusionRenderer~\cite{liang2025diffusionrenderer}. 
% However, these models typically demand substantially more training data, computational resources, and operate at lower resolutions, making them difficult to apply to driving scenes directly.
% Building a temporally consistent framework is an important direction for future work, but it remains beyond the scope of this paper.
}

\section{Conclusion}
We propose IntrinsicWeather, a novel framework for controllable weather editing in intrinsic space. Our approach achieves robust intrinsic decomposition across diverse weather and illumination conditions while enabling controlled weather editing based on maps and prompts. For the inverse renderer, we propose IMAA to provide attention guidance to help the model focus on semantically important regions. For the forward renderer, we leverage CLIP interpolation and diffusion priors to achieve fine-grained weather control. Last, we construct two datasets, WeatherSynthetic and WeatherReal, containing intrinsic maps to address the lack of large-scale autonomous driving rendering datasets under varied weather conditions. \added{Our IntrinsicWeather demonstrates performing weather editing in intrinsic space can achieve controllable and plausible editing while preserving geometry and material. Future work includes extending the framework to video-based weather editing and collecting more diverse and realistic data.}     

% future work
% \revise{
% Future directions include reducing dependence on high-quality training data. Combining diffusion models with reinforcement learning—guided by human or LLM feedback—could enhance robustness. Moreover, the emergence of auto-regressive generative models presents opportunities to further advance both rendering and inverse rendering tasks.
% }
\section*{Acknowledgement}
We thank the reviewers for the valuable comments. This work has been partially supported by the National Natural Science Foundation of China under grant No. 62572230.

{
    \small
    \bibliographystyle{ieeenat_fullname}
    \bibliography{main}
}

% \input{sec/8_suppl}

\end{document}